%% file: emnlp2022.tex
\pdfoutput=1

\documentclass[11pt]{article}

\usepackage{EMNLP2022}

\usepackage{times}
\usepackage{latexsym}

\usepackage[T1]{fontenc}

\usepackage[utf8]{inputenc}

\usepackage{microtype}

\usepackage{inconsolata}

\usepackage{url}
\usepackage{amsmath}
\usepackage{booktabs}
\usepackage{multirow}
\usepackage{makecell}
\usepackage{graphicx}
\usepackage{tikz}
\usepackage{pgfplots}
\usepackage{enumitem}

\usetikzlibrary{shadows}
\usetikzlibrary{plotmarks}
\usetikzlibrary{patterns}
\usetikzlibrary{pgfplots.groupplots}
\usetikzlibrary{arrows}
\usetikzlibrary{decorations}
\usetikzlibrary{decorations.pathreplacing}
\usetikzlibrary{backgrounds}
\usetikzlibrary{fit}
\usetikzlibrary{positioning}
\usetikzlibrary{calc}
\usetikzlibrary{shapes.multipart}

\definecolor{ugreen}{cmyk}{1,0,1,0.498}
\definecolor{lyyblue}{cmyk}{0.8278,0.3333,0,0.2941}
\definecolor{lyygreen}{cmyk}{0.6813,0,0.725,0.3725}
\definecolor{lyyred}{cmyk}{0,0.8855,0.8767,0.1098}
\definecolor{dblue}{cmyk}{1,0.5487,0,0.5569}

\title{Eliciting Knowledge from Large Pre-Trained Models for Unsupervised Knowledge-Grounded Conversation}

\author{Yanyang Li, Jianqiao Zhao, Michael R. Lyu, Liwei Wang\thanks{\ \ Corresponding author.} \\
Department of Computer Science and Engineering, The Chinese University of Hong Kong \\
\texttt{\{yyli21,jqzhao,lyu,lwwang\}@cse.cuhk.edu.hk}\\
}

\begin{document}
\maketitle
\begin{abstract}

Recent advances in large-scale pre-training provide large models with the potential to learn knowledge from the raw text.
It is thus natural to ask whether it is possible to leverage these large models as knowledge bases for downstream tasks.
In this work, we answer the aforementioned question in unsupervised knowledge-grounded conversation.
We explore various methods that best elicit knowledge from large models.
Our human study indicates that, though hallucinations exist, large models post the unique advantage of being able to output common sense and summarize facts that cannot be directly retrieved from the search engine.
To better exploit such generated knowledge in dialogue generation, we treat the generated knowledge as a noisy knowledge source and propose the posterior-based reweighing as well as the noisy training strategy.
Empirical results on two benchmarks show advantages over the state-of-the-art methods.
\end{abstract}

\section{Introduction}

Knowledge-grounded conversation \cite{DBLP:conf/iclr/DinanRSFAW19,moghe-etal-2018-towards} is the task where the model could reply to a dialogue history based on extra knowledge.
Compared to standard conversational modeling, this extra knowledge prevents the model from generating generic and non-informative responses \cite{li-etal-2016-diversity}.
Typically, at each turn of the conversation, a pool of knowledge candidates will be retrieved from a knowledge base like unstructured documents (e.g., Wikipedia) \cite{DBLP:conf/iclr/DinanRSFAW19} or a structured knowledge graph \cite{dziri-etal-2021-neural}.
The model then learns to select the most related knowledge from this pool, in an unsupervised manner, to generate its response.

However, constructing and maintaining knowledge bases are time-consuming and expensive.
Recent studies have shown that large pre-trained models are capable of grasping knowledge from unsupervised text corpora and memorizing facts to their weights \cite{petroni-etal-2019-language,roberts-etal-2020-much,lewis-etal-2021-question,wang-etal-2021-generative,liu-etal-2022-generated}.
These large models can even perform reasoning implicitly \cite{DBLP:journals/corr/abs-2201-11903}.
In light of this remarkable capacity of large models, we explore the possibility of leveraging large models as a new knowledge source for unsupervised knowledge-grounded conversation.

In this work, we first investigate the quality of knowledge generated by large models.
We examine which tuning method, including the conventional fine-tuning \cite{devlin-etal-2019-bert,DBLP:conf/iclr/0007WKWA21} and the recently proposed prefix-tuning \cite{li-liang-2021-prefix}, best prompts knowledge from large models for a given dialogue history.
We then design a human evaluation protocol and conduct an extensive quality assessment of the generated knowledge.
Despite some extent of hallucinations (plausible statements with factual errors) persist \cite{maynez-etal-2020-faithfulness}, large models can mostly generate related and correct knowledge for the future development of dialogue.
Moreover, some of this knowledge is not simply paraphrased or copied from web pages: they summarize scattered facts on the Internet (See Section \ref{sec:study}).
These observations advocate the unique value of employing large models as knowledge bases.

Owing to the hallucinations, it is risky to put generated knowledge directly into the dialogue system as the misinformation could contaminate the response.
We instead consider the generated knowledge as a noisy knowledge source and use it to aid the knowledge selection process.
Specifically, we measure its similarity to each knowledge candidate and refine the knowledge selection accordingly (See Section \ref{sec:reweigh}).
We further estimate the posterior of the refined knowledge selection distribution, inspired by the fact that the posterior detangles the one-to-many relation between dialogue context and knowledge selection \cite{DBLP:conf/iclr/KimAK20}.
In addition, we propose a noisy training strategy to strengthen the model's ability on handling noisy knowledge (See Section \ref{sec:train}).
All these strategies significantly elevate the performance of the existing state-of-the-art model to a new level on two widely-adopted benchmarks, Wizard of Wikipedia \cite{DBLP:conf/iclr/DinanRSFAW19} and Holl-E \cite{moghe-etal-2018-towards}.

\section{Eliciting Knowledge from Large Models}

In this section, we first introduce the tuning methods and large pre-trained models we used to generate knowledge for a given dialogue history.
Then we show the tagset developed for evaluating the generated knowledge.

\subsection{Methods and Models}
\label{sec:know-gen}

\begin{figure}
    \centering
    \includegraphics[scale=0.13]{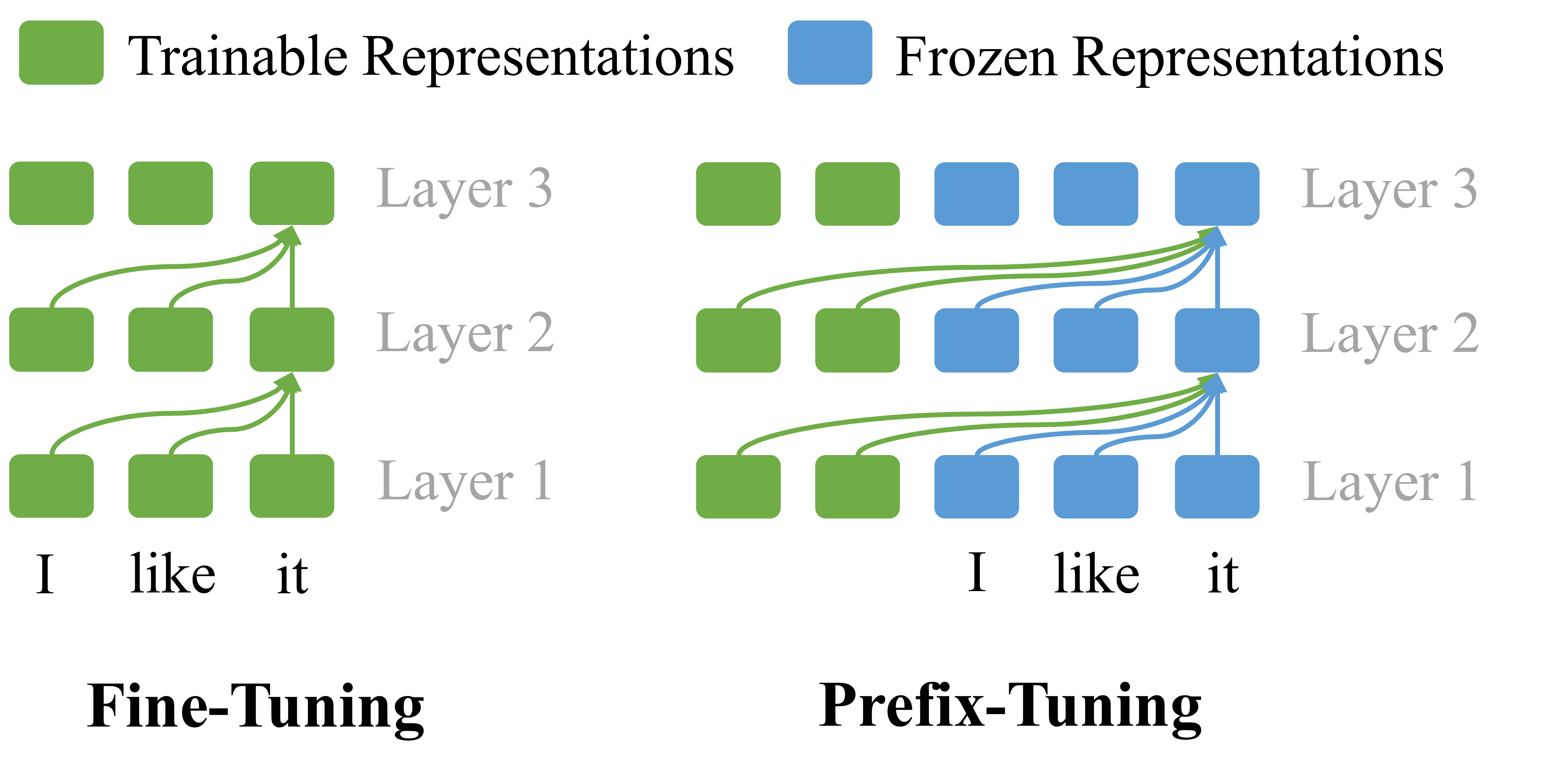}
    \caption{Comparison of fine-tuning \cite{devlin-etal-2019-bert} and prefix-tuning \cite{li-liang-2021-prefix}.}
    \label{fig:tuning}
    \vspace{-0.4cm}
\end{figure}

Since the objective function of large pre-trained models is to predict possible words instead of knowledge given the context \cite{devlin-etal-2019-bert,radford2019language}, tuning these large models on annotated data is necessary.
Here we focus on two tuning methods, as shown in Figure \ref{fig:tuning}:
\begin{itemize}[noitemsep,nolistsep]
    \item \textbf{Fine-Tuning} \cite{devlin-etal-2019-bert,DBLP:conf/iclr/0007WKWA21} which updates all weights in the model.
    \item \textbf{Prefix-Tuning} \cite{li-liang-2021-prefix} which freezes the pre-trained weights and tunes only a small set of parameters that are added as the prefix of the model's input.
\end{itemize}
Fine-tuning remains the standard approach for leveraging pre-trained models in downstream tasks, while prefix-tuning has a comparable performance but avoids the risk of catastrophic forgetting \cite{goodfellow2013empirical}, which is desirable in our task.

Another challenge is selecting large models.
Because our task requires large models to understand the dialogue history and then recommend a related knowledge piece for the user to follow up, we study two types of large models:
\begin{itemize}[noitemsep,nolistsep]
    \item \textbf{Pre-trained Language Models} (PLMs) that are trained on web documents with access to abundant knowledge during pre-training.
    \item \textbf{Pre-trained Dialogue Models} (PDMs) that are trained on dialogue data to better understand the dialogue history.
\end{itemize}
We choose T5 \cite{JMLR:v21:20-074} as the representative of PLMs and DialoGPT \cite{zhang-etal-2020-dialogpt} for PDMs, because they release a series of checkpoints with different model sizes.

Besides, we experiment with various decoding methods to see which of them best suits each type of large models, including greedy decoding, beam search and top-K sampling \cite{fan-etal-2018-hierarchical}.
We find that PDMs work best with top-K sampling and beam search for PLMs.

\subsection{Annotation Tagset}
\label{sec:tagset}

\begin{table*}[t!]
    \centering
    {\small
    \begin{tabular}{r|l|p{10.5cm}}
        \toprule
        &
        \makecell[c]{\textbf{Tag}} & \makecell[c]{\textbf{Definition}} \\
        \midrule
        \multicolumn{1}{r}{} & \multicolumn{2}{l}{\emph{Context Understanding:}} \\
        1 & \texttt{Related} & The generated output discusses facts that are related to the conversation. \\
        2 & \texttt{Unrelated} & The generated output does not discuss facts that are related to the conversation. \\
        \midrule
        \multicolumn{1}{r}{} & \multicolumn{2}{l}{\emph{Tuning Effectiveness:}} \\
        3 & \texttt{Non-Verifiable} & The generated output does not contain facts that could be verified. \\
        4 & \texttt{Verifiable} & The generated output contains facts that could be verified. \\
        \midrule
        \multicolumn{1}{r}{} & \multicolumn{2}{l}{\emph{Fact-Checking:}} \\
        5 & \texttt{Supported} & One can find evidence from the knowledge base to validate the factual information in the generated output. \\
        6 & \quad\texttt{Explicit Supported} & One only needs to find one evidence from the knowledge base for validation. \\
        7 & \quad\texttt{Implicit Supported} & One needs to find multiple evidences from the knowledge base for validation. \\
        8 & \texttt{Refuted} & One can find evidence from the knowledge base to contradict the factual information in the generated output. \\
        9 & \texttt{\textbf{N}ot \textbf{E}nough \textbf{I}nformation} & The factual information in the generated output could not be validated. \\
        10 & \quad\texttt{Reasonable NEI} & Though not validated by the knowledge base, the factual information matches common sense. \\
        11 & \quad\texttt{Unreasonable NEI} & Though not validated by the knowledge base, the factual information does not match common sense. \\
        12 & \quad\texttt{Hard NEI} & The factual information could not be validated by either the knowledge base or common sense. \\
        \bottomrule
    \end{tabular}
    }
    \caption{The tagset developed to evaluate the quality of the generated knowledge by human annotators.}
    \label{tab:tagset}
    \vspace{-0.4cm}
\end{table*}

To assess the quality of the generated knowledge, we develop an annotation tagset for human evaluation in Table \ref{tab:tagset}.
Each generated knowledge along with its associated dialogue history will be annotated by at least two tags, each from a different category: \emph{context understanding}, \emph{tuning effectiveness} (and \emph{fact-checking} if outputs contain facts).

\noindent\textbf{Context Understanding}
\texttt{Related} and \texttt{Unrelated} in rows 1-2 of Table \ref{tab:tagset} measure whether large pre-trained models understand the conversation and generate related knowledge.
Although we can use automatic metrics like the F1 score that measures the distance between the generated knowledge and the ground truth knowledge as an alternative, a single reference only captures one possible future direction of the dialogue.
In this sense, human evaluation provides a more comprehensive assessment.

\noindent\textbf{Tuning Effectiveness}
\texttt{Non-Verifiable} (e.g., chitchat) and \texttt{Verifiable} in rows 3-4 indicate the reliability of the tuning methods for eliciting knowledge from large models.
If a tuning method is effective, models should generate outputs that contain \texttt{Verifiable} facts.

\noindent\textbf{Fact-Checking}
Among those \texttt{Verifiable} outputs, we classify them into \texttt{Supported} (facts is supported by evidence), \texttt{Refuted} (facts is refuted by evidence) and \texttt{Not Enough Information} (NEI, evidence is not found), as shown in rows 5-12 of Table \ref{tab:tagset}.
These tags are mainly adapted from \citet{gupta-etal-2022-dialfact}.
Annotators will gather trustworthy evidence via search engines to determine the label.

To better understand the detailed behavior of large models, we divide \texttt{Supported} into \texttt{Explicit Supported} and \texttt{Implicit Supported}.
The former means that large models memorize existing documents, while the latter implies that they do more than memorization, e.g., summarization.
We also let the annotators check whether \texttt{NEI} outputs could be validated by common sense.
If common sense could be used for validation, these \texttt{NEI} outputs will be further classified into \texttt{Reasonable NEI} (facts match common sense) or \texttt{Unreasonable NEI} (facts contradict common sense), and \texttt{Hard NEI} if common sense is not applicable.

\section{Exploiting Generated Knowledge for Conversation}

In this section, we first review the state-of-the-art approach - PLATO-KAG \cite{huang-etal-2021-plato}.
Then we develop our method on top of PLATO-KAG to exploit generated knowledge.

\begin{figure*}
    \centering
    \includegraphics[scale=0.13]{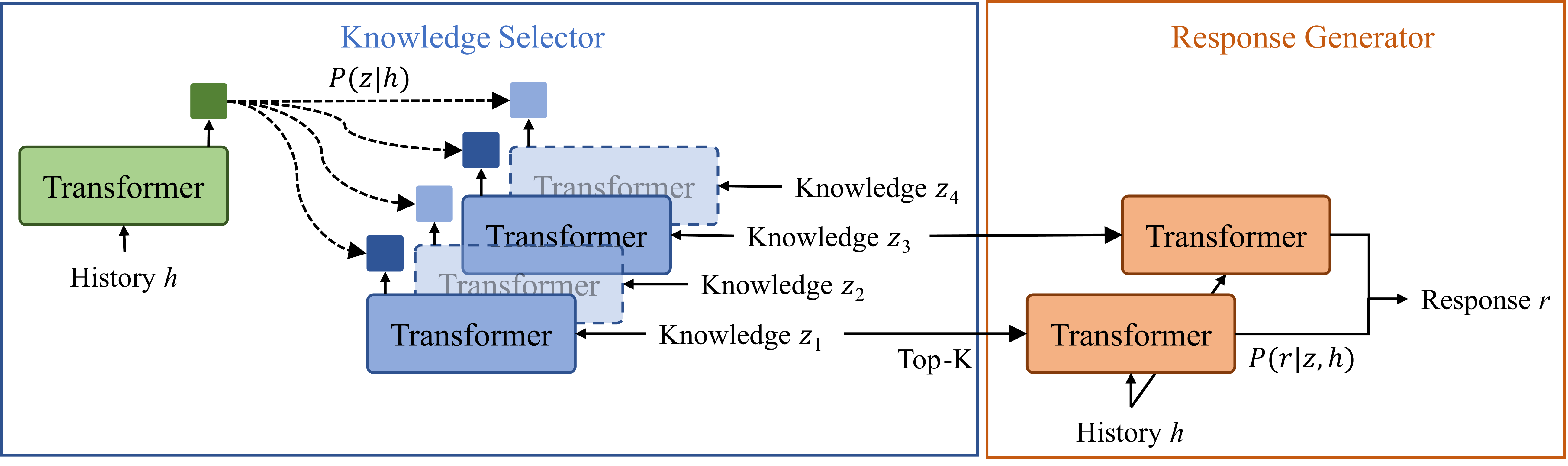}
    \caption{Illustration of PLATO-KAG \cite{huang-etal-2021-plato}.}
    \label{fig:plato-kag}
    \vspace{-0.4cm}
\end{figure*}

\subsection{PLATO-KAG}

As shown in Figure \ref{fig:plato-kag}, PLATO-KAG is a model consisting of two modules: a knowledge selector which selects top-K most relevant knowledge to the dialogue history from a pool of retrieved knowledge candidates, and a response generator that generates the response based on the dialogue history and the selected knowledge.

\noindent\textbf{Knowledge Selector}
The knowledge selector adopts a dual encoder with shared parameters to extract features.
The dialogue history $h$ and a knowledge candidate $z$ will pass to this encoder independently to get their own representations.
Then it estimates the relevance between the dialogue history $h$ and a knowledge candidate $z$ by:
\begin{equation}
    f(h,z)=(W_hE(h))^T(W_zE(z))
    \label{eqn:sim}
\end{equation}
where $E(\cdot)$ is the fixed-length vector representation of the input, i.e., the encoder's output on the [CLS] token.
$W_h$ and $W_z$ are two linear projections.

To select the top-K knowledge candidates, the knowledge selector computes the relevance between $h$ and all possible $z$.
Then only the top-K most related knowledge $Z=\{z_1,\cdots,z_K\}$ is retained to construct the knowledge selection distribution $P(z|h)$ as follows:
\begin{equation}
    P(z|h)=\frac{e^{f(h,z)}}{\sum_{z'\in Z}e^{f(h,z')}}
    \label{eqn:select}
\end{equation}

\noindent\textbf{Response Generator}
After the knowledge selection, the response generator will predict the probability of the response $r$ by:
\begin{equation}
    P(r|h)=\sum_{z\in Z}P(z|h)P(r|h,z)
\end{equation}
where $P(r|h,z)=\prod_i P(r_i|h,z,r_{<i})$ is a decoder that generates response $r$ given the dialogue history $h$ and one knowledge candidate $z$.

\subsection{Posterior-based Reweighing}
\label{sec:reweigh}

\noindent\textbf{Reweighing}
Generated knowledge contains hallucinated facts, as later shown in Section \ref{sec:human}.
It is thus not viable to take generated knowledge $g$ as the direct input of the model.
Instead, we interpret $g$ as noisy ground truth and define a refined knowledge selection distribution $P^*(z|h)$:
\begin{equation}
    \label{eqn:refine}
    P^*(z|h)=P(z|h)P(z|g)
\end{equation}
where $P(z|h)$ is the original knowledge selection distribution and $P(z|g)$ measures the similarity between $g$ and $z$.
This refined distribution $P^*(z|h)$ will score $z$ high only if $z$ is close to the history $h$ as well as the noisy ground truth $g$.

Similar to Eqs. \ref{eqn:sim}-\ref{eqn:select}, we define $P(z|g)$ to measure the closeness between $g$ and each $z$:
\begin{align}
    f(g,z)&=(W_hE(g))^T(W_zE(z)) \\
    P(z|g)&=\frac{e^{f(g,z)/\alpha}}{\sum_{z'\in Z}e^{f(g,z')/\alpha}}
\end{align}
where $\alpha$ is a hyper-parameter that controls the sharpness of $P(z|g)$.

\noindent\textbf{Posterior}
\citet{DBLP:conf/iclr/KimAK20} suggests that the posterior selection distribution $P(z|h,r)$ could select a more appropriate knowledge than the prior selection distribution $P(z|h)$, because the presence of future response $r$ narrows down the scope of all possible $z$.
We drive the posterior of the refined selection distribution $P^*(z|h)$ via the Bayes rule:
\begin{equation}
    P(z|h,r)=\frac{P(r|h,z)P^*(z|h)}{\sum_{z'\in Z} P(r|h,z')P^*(z'|h)}
    \label{eqn:posterior}
\end{equation}
where the denominator is tractable as there are only a small number of $z$ and $P(r|h,z)$ is exactly the response generator.

The main challenge is how to estimate $P(r|h,z)$ when $r$ is not yet observed.
We let the response generator greedy decode a most likely response $\bar{r}$ for a given $z$.
However, different $z$ might result in $\bar{r}$ with various lengths.
A long $\bar{r}$ tends to have a lower probability and is not competitive with the short one \cite{yang-etal-2018-breaking}.
We therefore use the mean token probability as the approximation of $P(r|h,z)\approx \frac{1}{N}\sum^N_{i=1}P(\bar{r}_i|h,z,\bar{r}_{<i})$, where $N$ is the length of $\bar{r}$.

In the end, we add a hyper-parameter $\beta$ to control the sharpness of the posterior $P^*(z|h,r)\propto P(z|h,r)^\beta$.
Since we only apply the Bayes rule once to obtain the posterior, adjusting the sharpness help to amplify or diminish the impact of updating.

\subsection{Noisy Training}
\label{sec:train}

Note that Eq. \ref{eqn:posterior} is only applicable when the response generator $P(r|h,z)$ is able to denoise, i.e., the model should give the likelihood estimate of a low value if $z$ is not appropriate.
In this case, the Bayes rule will update the posterior by lowering the chance of this inappropriate $z$ being selected.
Since the knowledge selector always presents its most confident selection to the response generator and the knowledge selector performs much better in the training set (a top-K accuracy of 90.3\% for the baseline) than in the test set (a top-K accuracy of 68.1\%),  such a bias will lead to a training-inference discrepancy and therefore the response generator is not resilience to noisy knowledge during testing.

To alleviate this issue, we employ the Gumbel-TopK trick \cite{DBLP:conf/icml/KoolHW19}, which adds noise to the top-K operation in the knowledge selector during training.
Specifically, we sample noise from the Gumbel distribution with location $\mu=0$ and scale $\phi=1$.
This noise will add to $f(h,z)$ in Eq. \ref{eqn:sim} to permute the ranking of knowledge candidates and perturb the selection distribution $P(z|h)$.

\section{Experimental Setup}

\subsection{Datasets}
\label{sec:setup}

We conduct experiments on two popular benchmarks: Wizard of Wikipedia \cite{DBLP:conf/iclr/DinanRSFAW19} (WoW), and Holl-E \cite{moghe-etal-2018-towards}.
The WoW dataset covers a wide range of topics (1,365 in total).
Each conversation in WoW happens between a wizard who has access to knowledge from Wikipedia about a specific topic and an apprentice who learns from the wizard about the topic.
Specifically, for our knowledge generation task in Section~\ref{sec:know-gen}, the input is the dialogue history and the target is the ground truth knowledge that the wizard used to generate his response.
There are 18,340/1,948/1,933 dialogues in the training/validation/test set.
The validation and test sets are split into two categories: \emph{Seen} which contains new dialogues with topics that appeared in the training set and \emph{Unseen} whose dialogues have topics that never appear in the training set.
We follow \citet{DBLP:conf/iclr/DinanRSFAW19}'s scripts to preprocess the data.

Compared to WoW, conversations in Holl-E happened between two participants discussing a specific movie, where a single document about that movie is given as knowledge.
There are 7,228/930/913 dialogues in the training/validation/test split.
We follow \citet{DBLP:conf/iclr/KimAK20}'s scripts for data preprocessing.

\subsection{Evaluation Metrics}

We assess all results (generated knowledge and responses) via both the automatic metric and human evaluation.

\noindent\textbf{Knowledge Generation Assessment}
In automatic evaluation, we compute the unigram F1 between the generated knowledge and the ground truth knowledge.
In human evaluation, we recruit three well-trained annotators who are fluent in English to evaluate 100 random samples from the seen and unseen test sets each, according to the scheme we proposed in Section \ref{sec:tagset}.
The tag of an example is determined by the majority vote of the three annotators.
The agreement among the annotators is measured via Fleiss’ kappa \cite{fleiss1971measuring}.

\noindent\textbf{Response Generation Assessment}
In the automatic evaluation, we report the perplexity (PPL) and Unigram F1 of ground truth responses.
We also collect the top-1 knowledge accuracy (P@1) statistics, which evaluate the performance of the knowledge selector.
In the human evaluation, 100 random examples from WoW seen and unseen test sets are distributed to three annotators respectively.
They will evaluate these samples in four aspects, following \citet{huang-etal-2021-plato}:
\begin{itemize}[noitemsep,nolistsep]
    \item \textbf{Coherent} measures whether the response is consistent with the dialogue history.
    \item \textbf{Informativeness} evaluates whether the response is generic and non-informative or not.
    \item \textbf{Engagingness} assesses how likely the annotator is willing to continue the discussion.
    \item \textbf{Hallucination} checks the correctness of the contained factual information.
\end{itemize}
Coherence, informativeness and engagingness are in the range of [0, 1, 2].
A higher value implies a better result.
Hallucination is in the range of [0, 1], where 0 means the response is factually correct and 1 means the response contains hallucinated facts.
We refer the readers to \citet{huang-etal-2021-plato} for more details.
The final score of each sample is determined through majority voting.

\subsection{Response Generation Baselines}

\noindent\textbf{TMN}
is the baseline released along with the WoW dataset \cite{DBLP:conf/iclr/DinanRSFAW19}.
It stores knowledge candidates' features in the memory for selection.
We include the released unsupervised trained checkpoint in our experiments\footnote{\url{https://parl.ai/projects/wizard_of_wikipedia}}.

\noindent\textbf{SKT}
models the knowledge selection process in multi-turn dialogue generation as a sequential latent variable model
\cite{DBLP:conf/iclr/KimAK20}.
We use their open-sourced models in our experiments\footnote{\url{https://github.com/bckim92/sequential-knowledge-transformer}}.

\noindent\textbf{KnowledGPT}
fine-tunes a GPT-2 \cite{radford2019language} and leverages the reinforcement learning approach to train an unsupervised sequential knowledge selector
\cite{zhao-etal-2020-knowledge-grounded}.
We adopt their released model for experiments\footnote{\url{https://github.com/zhaoxlpku/KnowledGPT}}.

\begin{table*}[t!]
    \centering
    {\small
    \begin{tabular}{clr|cc|cc}
        \toprule
        \multirow{2}*{\makecell[c]{\textbf{PLM}\\\textbf{Type}}} & \makecell[c]{\multirow{2}*{\textbf{Model}}} & \makecell[c]{\multirow{2}*{\textbf{\#Params}}} & \multicolumn{2}{c|}{\textbf{Fine-Tuning}} & \multicolumn{2}{c}{\textbf{Prefix-Tuning}} \\
        \cmidrule{4-7}
        & & & \textbf{Test Seen} & \textbf{Test Unseen} & \textbf{Test Seen} & \textbf{Test Unseen} \\
        \midrule
        N/A & T5-large & 737M & 0.2530 & 0.1044 & 0.1312 & 0.1245 \\
        \midrule
        \multirow{4}*{\rotatebox{90}{PLM}} & T5-small & 60M & 0.2521 & 0.1796 & 0.2138 & 0.1720 \\
        & T5-base & 222M & 0.2679 & 0.1807 & 0.2494 & 0.1735 \\
        & T5-large & 737M & 0.2624 & 0.1943 & 0.2575 & 0.1579  \\
        & T5-XL & 3B & 0.2684 & 0.2053 & 0.2629 & 0.1808 \\
        & T5-XXL & 11B & - & - & 0.2652 & 0.1874 \\
        \midrule
        \multirow{3}*{\rotatebox{90}{PDM}} & DialoGPT-small & 124M & 0.2357 & 0.1588 & 0.3041 & 0.1456 \\
        & DialoGPT-medium & 355M & 0.3216 & 0.1663 & 0.3173 & 0.1598 \\
        & DialoGPT-large & 774M & 0.3217 & 0.1705 & 0.3209 & 0.1613 \\
        \bottomrule
    \end{tabular}
    }
    \caption{The automatic evaluation result (Unigram F1) of two tuning methods for generating knowledge in WoW seen and unseen test sets (``N/A'' means we do not use the pre-trained weights). Results of fine-tuned T5-XXL are missing due to resource constraint.}
    \label{tab:wow-know-pred}
    \vspace{-0.4cm}
\end{table*}

\input{figures/knowledge}

\subsection{Implementation Details}

\noindent\textbf{Knowledge Generation}
For fine-tuning, all models use a batch size of 64, a learning rate of 5e-5, and the inverse square root learning rate scheduler \cite{DBLP:conf/nips/VaswaniSPUJGKP17} with 1000 warmup steps.
We validate the model on the validation set every 1000 steps and early stop the training if the performance does not improve after 15 validations.
For prefix-tuning, the prefix length is set to 5 as in \citet{li-liang-2021-prefix}.
Other hyper-parameters are almost the same as in fine-tuning, except that the learning rate is kept constant and reduced by 1/10 only if the validation set performance does not improve after 10 validations.
At inference, DialoGPT is decoded with top-K sampling where K is 10 and the beam size is 20.
For T5, we use beam search with a beam size of 10.

\noindent\textbf{Response Generation}
Since \citet{huang-etal-2021-plato} did not release their codes and models before we start the experiments, we reimplement their approach in ParlAI \cite{miller-etal-2017-parlai} and report our own results as well.
We follow \citet{huang-etal-2021-plato}'s hyper-parameters settings in our experiments.
For our proposed reweighing method, we perform a grid search on the validation set ($\alpha\in[1,10],\beta\in(0,1)$) and set $\alpha=5,\beta=0.4$.
According to Table \ref{tab:wow-know-pred}, we choose the generated knowledge of DialoGPT-large for our experiments, as it performs the best on average.
All experiments are conducted on 8 NVIDIA A100 80G.
It takes roughly one day to train one model.

\section{Results and Analysis}

\subsection{Knowledge Generation Results}
\label{sec:study}

We conduct a case study of eliciting knowledge from large models on the WoW dataset and present the evaluation results.

\subsubsection{Automatic Evaluation Results}

Table \ref{tab:wow-know-pred} shows F1 scores of various large models tuned by different methods on the seen and unseen test sets.
Results of fine-tuning T5-XXL are missing because we do not have enough resources to train this model.
The first row of Table \ref{tab:wow-know-pred} is the baseline result of tuning a randomly initialized T5-large model.
We observe that nearly all large models perform better than this baseline, especially on the unseen test set.
This observation indicates the per-trained weights do store a lot of factual information as they make a non-trivial improvement.

We also see that PDMs perform much better than PLMs on data with a seen topic, while PLMs are better on the unseen topic in most cases.
This might be the consequence that PLMs are trained on diverse text data, which allows them to generalize better on unseen topics.
PDMs, on the other hand, are trained on dialogue data only and have a smaller discrepancy between pre-training and fine-tuning.
Thus PDMs perform better on seen topics.
We also find that the results of fine-tuning are much better than prefix-tuning in general.
But this gap is closed when the model gets larger, which is aligned with the conclusion in \citet{lester-etal-2021-power}.

Interestingly, large models scale poorly on our task.
On the unseen test set, the performance increases only around 3 points while the model size is 50$\times$ larger (0.1796 for fine-tuned T5-small with 60M parameters vs. 0.2053 for fine-tuned T5-XL with 3B parameters).

\begin{table*}[t!]
    \centering
    \setlength{\tabcolsep}{4pt}
    {\small
    \begin{tabular}{l|ccc|cccc}
        \toprule
        \makecell[c]{\multirow{3}*{\textbf{Test Seen}}} & \multicolumn{3}{c|}{\textbf{Automatic Evaluation}} & \multicolumn{4}{c}{\textbf{Human Evaluation}} \\
        \cmidrule{2-8}
        & \textbf{PPL$\Downarrow$} & \textbf{P@1$\Uparrow$} & \textbf{Unigram F1$\Uparrow$} & \textbf{Coherence$\Uparrow$} & \textbf{Informativeness$\Uparrow$} & \textbf{Engagingness$\Uparrow$} & \textbf{Hallucination$\Downarrow$} \\
        \midrule
        TMN & 61.21 & 0.220 & 0.172 & 0.4757 & 0.3883 & 0.4175 & 0.0777 \\
        SKT & 57.27 & 0.258 & 0.187 & 0.9806 & 0.7767 & 0.6990 & 0.0680 \\
        KnowledGPT & 19.60 & 0.262 & \underline{0.209} & 1.0000 & \underline{\textbf{1.2330}} & 1.0874 & \underline{\textbf{0.0097}} \\
        PLATO-KAG & 9.767 & 0.253 & 0.188 & - & - & - & - \\
        \midrule
        PLATO-KAG$^*$ & 11.51 & \underline{\textbf{0.266}} & 0.207 & \underline{1.4757} & 1.1748 & \underline{\textbf{1.2816}} & 0.0388 \\
        PLATO-KAG$^+$ & 12.37 & 0.254 & \textbf{0.211} & \textbf{1.4951} & 1.1845 & 1.2718 & 0.0291 \\
        \bottomrule
        \toprule
        \makecell[c]{\multirow{3}*{\textbf{Test Unseen}}} & \multicolumn{3}{c|}{\textbf{Automatic Evaluation}} & \multicolumn{4}{c}{\textbf{Human Evaluation}} \\
        \cmidrule{2-8}
        & \textbf{PPL$\Downarrow$} & \textbf{P@1$\Uparrow$} & \textbf{Unigram F1$\Uparrow$} & \textbf{Coherence$\Uparrow$} & \textbf{Informativeness$\Uparrow$} & \textbf{Engagingness$\Uparrow$} & \textbf{Hallucination$\Downarrow$} \\
        \midrule
        TMN & 103.1 & 0.112 & 0.150 & 0.5000 & 0.2788 & 0.3173 & 0.1058 \\
        SKT & 87.93 & 0.177 & 0.157 & 0.7019 & 0.5000 & 0.5385 & 0.0673 \\
        KnowledGPT & 22.85 & 0.238 & \underline{0.196} & 0.9712 & 0.9904 & 0.7692 & \underline{\textbf{0.0096}} \\
        PLATO-KAG & 11.46 & \underline{\textbf{0.253}} & 0.181 & - & - & - & - \\
        \midrule
        PLATO-KAG$^*$ & 12.75 & 0.233 & \underline{0.196} & \underline{\textbf{1.4327}} & \underline{\textbf{1.2019}} & \underline{\textbf{1.2019}} & 0.0962 \\
        PLATO-KAG$^+$ & 13.77 & 0.231 & \textbf{0.203} & 1.2596 & 1.0192 & 1.0096 & 0.0385 \\
        \bottomrule
    \end{tabular}
    }
    \caption{The automatic and human evaluation results on WoW seen (upper) and unseen (bottom) test sets. $^*$ means this is our implementation results. $^+$ means our proposed method is applied. Note that PPL is generally not comparable among baselines, as their vocabularies are different. The best results are in \textbf{bold} and the best baseline results are \underline{underlined}.}
    \label{tab:wow}
    \vspace{-0.4cm}
\end{table*}

\subsubsection{Human Evaluation Results}
\label{sec:human}

The human evaluation results are presented in Figure \ref{fig:knowledge}.
This evaluation has a kappa value of 1 for the context understanding dimension and 0.698 for the remaining dimensions\footnote{Since \emph{fact-checking} is a fine-grained category of \texttt{Verifiable} in \emph{tuning effectiveness}, we merge these two categories and compute the kappa value jointly.}.
Here we study the outputs of fine-tuned DialoGPT-large, as it performs the best on average.
We also analyze prefix-tuned T5-XXL in Appendix \ref{sec:know-t5} and the results are similar.

\noindent\textbf{Context Understanding}
From the first subplot of Figure \ref{fig:knowledge}, large models can reliably generate related knowledge for a given dialogue history, where around 90\% tags are \texttt{Related} in both test sets.

\noindent\textbf{Tuning Effectiveness}
The second diagram in Figure \ref{fig:knowledge} shows that large models exhibit desirable behaviors after fine-tuning: it generates knowledge (\texttt{Verifiable}) in all cases.

\noindent\textbf{Fact-Checking}
The rightmost three panels of Figure \ref{fig:knowledge} demonstrate the factual correctness of the generated knowledge.
As shown in the third panel, large models generate factually correct (\texttt{Supported}) statements in most cases, though there is still around 10\% of the chance to produce hallucinated information (\texttt{Refuted}).
Among all the factually correct knowledge (the second last panel), more than 50\% of them are \texttt{Implicit Supported}.
This is exciting as large models are able to assemble multiple facts in their outputs, which cannot be substituted by simple search engine retrieval.
This ability to summarize justifies the value of large models in serving as knowledge bases.

The last panel of Figure \ref{fig:knowledge} checks whether \texttt{NEI} claims could be verified by common sense.
There is a certain amount (39$\sim$57\%) of \texttt{NEI} claims that are common sense.
This observation advocates another advantage of utilizing large models as knowledge bases: they can provide common-sense information that lies behind the human mind, with no need for humans to explicitly write them down.
We show some annotated examples in Appendix \ref{sec:know-example}.

\subsection{Response Generation Results}

\subsubsection{Main Results}

Table \ref{tab:wow} is the response generation results of the WoW test sets.
We can see that applying our proposed method to PLATO-KAG obtains the highest F1 score, even if our reimplemented PLATO-KAG baseline already performs much better than reported in the paper.
On the other hand, our proposed method seems to lower the top-1 knowledge accuracy, i.e., P@1 drops from 0.266 to 0.254 in the seen test set and from 0.233 to 0.231 in the unseen test set.
Note that PLATO-KAG is a model whose input consists of K knowledge candidates.
If the ground truth knowledge is not ranked in the first place but presented in the top-K results, the model can still use the ground truth for the generation.
In this case, the top-K knowledge accuracy is a more important metric for evaluating knowledge selection.
Though not presented in Table \ref{tab:wow}, P@K increases from 0.681 to 0.690 in the seen test set and from 0.645 to 0.656 in the unseen test set.
Table \ref{tab:holle} displays the automatic evaluation results in Holl-E datasets.
Similar to the results of WoW, our proposed method significantly outperforms the baseline systems in terms of the F1 score.

\begin{table}[t!]
    \centering
    {\small
    \begin{tabular}{l|ccc}
        \toprule
        \makecell[c]{\textbf{System}} & \textbf{PPL$\Downarrow$} & \textbf{P@1$\Uparrow$} & \textbf{Unigram F1$\Uparrow$} \\
        \midrule
        SKT & 52.02 & \underline{\textbf{0.303}} & 0.295 \\
        PLATO-KAG & 10.22 & 0.271 & 0.300 \\
        \midrule
        PLATO-KAG$^*$ & \underline{5.816} & 0.250 & \underline{0.310} \\
        PLATO-KAG$^+$ & \textbf{5.495} & 0.272 & \textbf{0.320} \\
        \bottomrule
    \end{tabular}
    }
    \caption{The evaluation results on Holl-E test set.}
    \label{tab:holle}
    \vspace{-0.4cm}
\end{table}

Table \ref{tab:wow} also reports the human evaluation results of WoW.
The kappa value of this human evaluation is 0.415.
In the seen test set, our strategy improves over baselines in nearly all metrics.
However, our method degrades the performance of the unseen test set.
In Section \ref{sec:analysis}, we will show that our method put significantly more ground truth knowledge into the responses.
In spite of that more knowledge helps to reduce hallucinations (from 0.0962 to 0.0385 as shown in Table \ref{tab:wow}), this could also lead to a degenerated result in human evaluation \cite{huang-etal-2021-plato}, as the knowledge makes the response far less interesting.

\subsubsection{Analysis}
\label{sec:analysis}

We conduct an ablation study in Table \ref{tab:wow-ablation} for a better understanding of our proposed method.
We additionally report Knowledge F1, the F1 score between the generated response and the ground truth knowledge \cite{DBLP:conf/ijcai/LianXWPW19,shuster-etal-2021-retrieval-augmentation}, which indicates how much ground truth knowledge is embedded into the response.

As shown in Table \ref{tab:wow-ablation}, all steps in our proposed method, including noisy training and posterior-based reweighing, contribute to the final performance.
In particular, reweighing greatly improves Knowledge F1, which implies that it helps to select and incorporate ground truth knowledge into the response generation.
We give some examples in Appendix \ref{sec:dial-example} for demonstration.

\section{Related Work}

\noindent\textbf{Knowledge-Grounded Conversation}
The dialogue system field has witnessed a growing interest in knowledge-grounded conversation in recent years.
Many related benchmarks have been proposed to study this problem \cite{zhang-etal-2018-personalizing,zhou-etal-2018-dataset,DBLP:conf/iclr/DinanRSFAW19,gopalakrishnan19_interspeech,komeili-etal-2022-internet}.
Early work \cite{DBLP:conf/iclr/DinanRSFAW19} had harnessed the annotated knowledge for training.
Unsupervised approaches become attractive as acquiring these annotations is expensive.
\citet{zhao-etal-2020-knowledge-grounded} use reinforcement learning to fine-tune GPT-2 \cite{radford2019language} for unsupervised knowledge selection.
\citet{huang-etal-2021-plato} achieve a new state-of-the-art by selecting top-K knowledge when annotations are not available.
Another line of research improves the knowledge selection modeling by estimating the posterior, which makes use of the future utterance.
\citet{DBLP:conf/ijcai/LianXWPW19} train the knowledge selector as a variational auto-encoder  \cite{DBLP:journals/corr/KingmaW13}.
\citet{DBLP:conf/iclr/KimAK20} further model the knowledge selection in multi-turn dialogue as a sequential latent variable.
More recently, dialogue model pre-training also attempts to involve knowledge for generating informative responses.
\citet{shuster-etal-2021-retrieval-augmentation} utilize the pre-trained retriever DPR \cite{karpukhin-etal-2020-dense}.
\citet{DBLP:journals/corr/abs-2201-08239} directly access to the search engine to collect relevant knowledge.

\begin{table}[t!]
    \centering
    \setlength{\tabcolsep}{3pt}
    {\small
    \begin{tabular}{l|cc|cc}
        \toprule
        \makecell[c]{\multirow{3}*{\textbf{System}}} & \multicolumn{2}{c|}{\textbf{Test Seen}} & \multicolumn{2}{c}{\textbf{Test Unseen}} \\
        \cmidrule{2-5}
        & \makecell[c]{\textbf{Unigram}\\\textbf{F1$\Uparrow$}} & \makecell[c]{\textbf{Know.}\\\textbf{F1$\Uparrow$}} & \makecell[c]{\textbf{Unigram}\\\textbf{F1$\Uparrow$}} & \makecell[c]{\textbf{Know.}\\\textbf{F1$\Uparrow$}} \\
        \midrule
        PLATO-KAG$^*$ & 0.208 & 0.193 & 0.196 & 0.183  \\
        + Noisy Train. & 0.209 & 0.192 & 0.203 & 0.188  \\
        + Post. Reweigh & 0.211 & 0.200 & 0.203 & 0.193 \\
        \bottomrule
    \end{tabular}
    }
    \caption{The ablation study on WoW test sets.}
    \label{tab:wow-ablation}
    \vspace{-0.4cm}
\end{table}

\noindent\textbf{Knowledge in Pre-Trained Models}
The LAMA prob \cite{petroni-etal-2019-language} first study knowledge stored in pre-trained models.
They show that pre-trained models contain a certain amount of factual knowledge without any fine-tuning.
This finding has motivated a series of work that adopts knowledge from pre-trained models for downstream tasks.
\citet{roberts-etal-2020-much} show that pre-trained models fine-tuned on question-answering datasets without accessing any external knowledge base could obtain a remarkable result.
\citet{DBLP:conf/kdd/WangQHLYWLSHSL22} prob relational structures from pre-trained models for Text-to-SQL parsing.
\citet{liu-etal-2022-generated} further demonstrate that pre-trained models can generate knowledge via prompting to help in common sense reasoning tasks.
Perhaps the most related work is \citet{liu-etal-2022-multi}, where they adapt a large model to knowledge-grounded conversation via multi-stage prompting and which includes an intermediate knowledge generation step.
Compared to this work, our work treats large models as a general-purpose knowledge base, then elicits and transfers knowledge from it to improve a small but strong downstream task model with a distinct architecture.

\noindent\textbf{Knowledge Distillation}
Our work also closely resembles knowledge distillation \cite{DBLP:journals/corr/HintonVD15}, as we similarly transfer knowledge from a large pre-trained model to a small downstream task model.
Most existing approaches employ continuous vectors to represent knowledge, e.g., logits \cite{DBLP:journals/corr/HintonVD15}, attention distribution \cite{DBLP:conf/nips/WangW0B0020}, hidden features \cite{DBLP:journals/corr/RomeroBKCGB14} or weights \cite{lin-etal-2021-weight}, which are not straightforwardly interpretable.
In this work, the large model generates discrete, readable sentences to transfer knowledge.

\section{Conclusion}

In this work, we show that large pre-trained models could serve as knowledge bases for unsupervised knowledge-grounded conversation.
The study on the generated knowledge of large models has the following observations:
\begin{itemize}[noitemsep,nolistsep]
    \item Fine-tuning better elicits knowledge from large models than prefix-tuning.
    \item Knowledge pieces generated by pre-trained language models have a higher quality on unseen topics, while those from pre-trained dialogue models are better on seen topics.
    \item Large pre-trained models can synthesize common sense and summarize facts scattered on the web.
\end{itemize}
We also propose posterior-based reweighing and noisy training, which helps to incorporate the generated knowledge into the dialogue system.
These simple strategies show a promising result over the strong baselines.

\section*{Limitations}

We realize that there still are some limitations in our work despite the exciting results:

\noindent\textbf{Generated Knowledge}
A certain amount of annotated data is required to fine-tune large models before they could output knowledge stored in their weights.
Such kind of data could be difficult to collect as it requires highly educated annotators.
Besides, although large models could generate knowledge that needs multiple pieces of evidence for verification (samples with \texttt{Implicit Supported} tag), to what extent large models understand facts and the relation to hallucination remain unknown.

\noindent\textbf{Proposed Method}
The approach we proposed to leverage the generated knowledge is still primitive, as it is purely training-free and applied only at inference.
In addition, we only use one generated knowledge sentence in experiments.
Aside from this, in human evaluation we have shown that injecting more knowledge into responses reduces hallucination, but results in the degradation of other dimensions like engagingness.
How to carefully balance these quality measurements is another topic that is worth investigating.

\section*{Acknowledgements}

This work is supported by Centre for Perceptual and Interactive Intelligence Limited, UGC under Research Matching Grant Scheme, and partially by Research Grants Council of the Hong Kong Special Administrative Region, China (No. CUHK 14210920 of the General Research Fund).


\bibliography{anthology,custom}
\bibliographystyle{acl_natbib}

\newpage

\appendix

\input{figures/knowledge-t5}

\section{Human Evaluation on T5-XXL}
\label{sec:know-t5}

Figure \ref{fig:knowledge-t5} is the human evaluation results on the generated knowledge of prefix-tuned T5-XXL.
The kappa value of the context understanding dimension is 0.922 and 0.726 for the remaining two dimensions.
Compared to DialoGPT-large, T5-XXL performs more robustly in generating related knowledge for unseen topics, as indicated by the leftmost panel of Figure \ref{fig:knowledge-t5}.
It also seems that prefix-tuning is slightly less effective, as it produces a few non-verifiable claims, as shown in the second leftmost subplot of Figure \ref{fig:knowledge-t5}.
In the rightmost diagram of Figure \ref{fig:knowledge-t5}, we note that T5-XXL is more likely to generate knowledge that is hard to verify.
This might indicate a higher risk of hallucination when using larger models.
Other measurements remain similar for DialoGPT-large and T5-XXL.

\section{Knowledge Generation Examples}
\label{sec:know-example}

Table \ref{tab:tag-example} gives some generated knowledge examples, each with a different tag annotated.
Most of these examples are generated by fine-tuned DialoGPT-large on WoW unseen test set.

\begin{table*}[t!]
    \centering
    {\small
    \begin{tabular}{c|l|p{12cm}}
        \toprule
        \multicolumn{2}{c|}{\textbf{Tag}} & \makecell[c]{\textbf{Example}} \\
        \midrule
        \multicolumn{2}{l|}{\multirow{4}*{\texttt{Related}}} & \makecell[c]{\emph{Dialogue History}} \\
        \multicolumn{2}{l|}{} & [Apprentice]: I love to bowl, but what is the games history, I wonder?\\
        \multicolumn{2}{l|}{} & \makecell[c]{\emph{Knowledge}} \\
        \multicolumn{2}{l|}{} & Bowling is an Olympic sport and is played at all levels of society and at all ages.\\
        \midrule
        \multicolumn{2}{l|}{\multirow{5}*[-0.3cm]{\texttt{Unrelated}}} & \makecell[c]{\emph{Dialogue History}} \\
        \multicolumn{2}{l|}{} & [Wizard]: I love the hunting game.More like a hunter\\
        \multicolumn{2}{l|}{} & [Apprentice]: What is the hunting game?  Tell me more about it.\\
        \cmidrule{3-3}
        \multicolumn{2}{l|}{} & \makecell[c]{\emph{Knowledge}} \\
        \multicolumn{2}{l|}{} & Hunting is the practice of killing or trapping animals, or pursuing or tracking them with the intent of doing so.\\
        \midrule
        \midrule
        \multicolumn{2}{l|}{\texttt{Non-Verifiable}} & The food is delicious. \\
        \midrule
        \multicolumn{2}{l|}{\texttt{Verifiable}} & American football, also known as American football or American football, is a team sport played between two teams of eleven players with a spherical ball. \\
        \midrule
        \midrule
        \multirow{2}*[-0.4cm]{\rotatebox{90}{\texttt{Supported}}} & \makecell[l]{\texttt{Explicit}\\\texttt{Supported}} & Dylan's Candy Bar is a chain of boutique candy shops and candy suppliers currently located in New York City; East Hampton, New York; Los Angeles, Chicago and Miami Beach, as well as in wholesale venues around the globe. \\
        \cmidrule{2-3}
        & \makecell[l]{\texttt{Implicit}\\\texttt{Supported}} & The Walking Dead is an American post-apocalyptic horror television series developed by Frank Darabont for AMC that is based on the comic book series of the same name by Robert Kirkman, Tony Moore, and Charlie Adlard. \\
        \midrule
        \multicolumn{2}{l|}{\texttt{Refuted}} & The dog was the first species to be domesticated and has been selectively bred over millennia for various behaviors, sensory capabilities, and physical attributes. \\
        \midrule
        \multirow{3}*[-0.4cm]{\rotatebox{90}{\texttt{NEI}}} & \makecell[l]{\texttt{Reasonable}\\\texttt{NEI}} & Bowling is an Olympic sport and is played at all levels of society and at all ages. \\
        \cmidrule{2-3}
        & \makecell[l]{\texttt{Unreasonable}\\\texttt{NEI}} & The sky is pale green. \\
        \cmidrule{2-3}
        & \texttt{Hard NEI} & The first stable line-up consisted of Michael "Mike D" Diamond (vocals, drums), Adam "MCA" Yauch (vocals, bass) and Adam "The Dope Man" Horovitz (vocals, guitar). \\
        \bottomrule
    \end{tabular}
    }
    \caption{Examples with different annotated tags (sampled from DialoGPT-large in WoW unseen test set except for \texttt{Non-Verifiable} and \texttt{Unreasonable NEI} as they do not appear in the data and thus are hand-crafted).}
    \label{tab:tag-example}
    \vspace{-0.4cm}
\end{table*}

\section{Response Generation Examples}
\label{sec:dial-example}

Table \ref{tab:dial-example} are response examples from different response generation models in two WoW test sets.
We can see that our PLATO-KAG$+$ generates responses that are more interesting than other baselines.
We also note that the first few words are similar for the PLATO-KAG baseline and our PLATO-KAG$+$.
It seems that our strategy improves the long-term modeling of PLATO-KAG to achieve a better result.

\begin{table*}[t!]
    \centering
    \setlength{\tabcolsep}{4pt}
    {\small
    \begin{tabular}{l|
                >{\raggedright\arraybackslash}
                p{\dimexpr.8\textwidth-\tabcolsep\relax}@{}}
        \toprule
        \multirow{5}*{\textbf{Context}}
        & [Apprentice]: I have been thinking about getting a poodle\\
        & [Wizard]: Oh yeah? You can choose from the Standard, Miniature, or Toy Poodles.\\
        & [Apprentice]: I think i might egt a toy\\
        & [Wizard]: Good selection. Toy poodles won "Best in Show" at Crufts in 1966 and 1982.\\
        & [Apprentice]: How big do they get? do you know\\
        \midrule
        TMN & they are the second most intelligent breed . \\
        SKT & i don ' t know , but i know they are a group of formal dog breeds . \\
        KnowledGPT & they are ranked second most intelligent dog breed just behind the border collie, the poodle is skillful in many dog sports and activities, including agility \\
        PLATO-KAG$^*$ & They are ranked as the second most intelligent dog breed. \\
        PLATO-KAG$^+$ & They are medium sized. They achieved the award in 1955, 1985, 2002 and 2014. \\
        \bottomrule
        \toprule
        \multirow{6}{*}[-1cm]{\textbf{Context}}
        & [Wizard]: Do you like Chihuahua's? I think they are cute, and their name origin is interesting to me, it is the smallest breed of dog\\
        & [Apprentice]: Yeah I love them! I own one! They are so friendly and nice!\\
        & [Wizard]: Dogs in general are so cute, it is crazy to me that they evolved from extant gray wolves, as wolves are so much different, it is hard to believe they are ancestors\\
        & [Apprentice]: It's amazing how humans domesticated and bred dogs down to create the dogs we have today. Early dogs were bread to ease everyday task.\\
        & [Wizard]: Yeah, while the origins of dogs are not really clear, it is clear humans had a role in that process, which amazes me, as we have created so many different and varying breeds of dogs\\
        & [Apprentice]: The Dochen was used back in the day to hunt out rabbits from holes in the ground for hunters to shoot\\
        \midrule
        TMN & i ' ve never had them , but i ' ve heard that they ' re so cute . i ' ve been reading about them since they ' re so cute . \\
        SKT & the earliest of these were found at tres adpots in mexico in veracruz , mexico in veracruz , mexico , mexico , and other animals . \\
        KnowledGPT & that's so cool, the dog was the first species to be domesticated and has been selectively bred over millennia for various behaviors, sensory capabilities, and \\
        PLATO-KAG$^*$ & That is so cool! I love dogs, they are so smart and friendly, and they are so smart and friendly with humans too! \\
        PLATO-KAG$^+$ & That's so cool! I didn't know that, I always thought they were just used for hunting, but I guess they were used for other things too. \\
        \bottomrule
    \end{tabular}
    }
    \caption{Examples of the generated responses on the WoW seen (upper) and unseen (bottom) test sets.}
    \label{tab:dial-example}
    \vspace{-0.4cm}
\end{table*}

\end{document}

%% file: figures/knowledge.tex
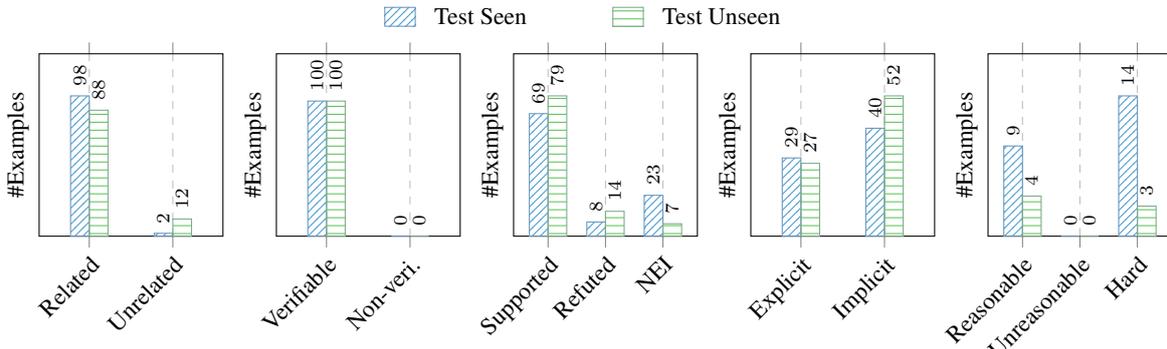
\begin{figure*}[t!]
    \centering
    \makeatletter
    \let\ref\@refstar
    \ref{grouplegend}
    \makeatother
    \vspace{-0.2in}
    \begin{tikzpicture}
        \begin{groupplot}[
            group style={group size=5 by 1, horizontal sep=20pt},
            width=1.0\textwidth,
            height=4cm,
            legend cell align={left},
            legend pos=north west,
            enlargelimits=0.1,
            legend style={
                font=\small,
                draw=none,
                column sep=5pt,
                /tikz/every even column/.append style={column sep=30pt},
                legend columns=2,
            },
        ]
        \nextgroupplot[
          width=0.25\linewidth,
          symbolic x coords={Related,Unrelated},
          enlarge x limits=0.6,
          enlarge y limits={upper,value=0.3},
          ylabel={\#Examples},
          ylabel near ticks,
          ybar=0pt,
          xtick=data,
          ytick=\empty,
          ymin=0,
          nodes near coords,
          every node near coord/.append style={rotate=90,anchor=west,font=\scriptsize},
          every tick label/.append style={font=\footnotesize},
          xticklabel style={rotate=45,anchor=north east,font=\footnotesize,inner sep=0pt,outer sep=2pt},
          bar width=7pt,
          xmajorgrids=true,
          ymajorgrids=true,
          grid style=dashed,
          label style={font=\small},
          legend to name=grouplegend,
          legend image code/.code={\draw [#1] (0cm,-0.1cm) rectangle ++(0.4cm,0.25cm);},
        ]
          \addplot [draw=lyyblue!60,fill=lyyblue!60,pattern=north east lines,pattern color=lyyblue!60] coordinates {(Related,98) (Unrelated,2)};
          \addlegendentry{Test Seen}
          \addplot [draw=lyygreen!60,fill=lyygreen!60,pattern=horizontal lines,pattern color=lyygreen!60] coordinates {(Related,88) (Unrelated,12)};
          \addlegendentry{Test Unseen}
        \nextgroupplot[
          width=0.25\linewidth,
          symbolic x coords={Verifiable,Non-veri.},
          enlarge x limits=0.6,
          enlarge y limits={upper,value=0.35},
          ylabel={\#Examples},
          ylabel near ticks,
          ybar=0pt,
          xtick=data,
          ytick=\empty,
          ymin=0,
          nodes near coords,
          every node near coord/.append style={rotate=90,anchor=west,font=\scriptsize},
          every tick label/.append style={font=\footnotesize},
          xticklabel style={rotate=45,anchor=north east,font=\footnotesize,inner sep=0pt,outer sep=2pt},
          bar width=7pt,
          xmajorgrids=true,
          ymajorgrids=true,
          grid style=dashed,
          label style={font=\small},
        ]
          \addplot [draw=lyyblue!60,fill=lyyblue!60,pattern=north east lines,pattern color=lyyblue!60] coordinates {(Verifiable,100) (Non-veri.,0)};
          \addplot [draw=lyygreen!60,fill=lyygreen!60,pattern=horizontal lines,pattern color=lyygreen!60] coordinates {(Verifiable,100) (Non-veri.,0)};
        \nextgroupplot[
          width=0.25\linewidth,
          symbolic x coords={Supported,Refuted,NEI},
          enlarge x limits=0.3,
          enlarge y limits={upper,value=0.3},
          ylabel={\#Examples},
          ylabel near ticks,
          ybar=0pt,
          xtick=data,
          ytick=\empty,
          ymin=0,
          nodes near coords,
          every node near coord/.append style={rotate=90,anchor=west,font=\scriptsize},
          every tick label/.append style={font=\footnotesize},
          xticklabel style={rotate=45,anchor=north east,font=\footnotesize,inner sep=0pt,outer sep=2pt},
          bar width=7pt,
          xmajorgrids=true,
          ymajorgrids=true,
          grid style=dashed,
          label style={font=\small},
        ]
          \addplot [draw=lyyblue!60,fill=lyyblue!60,pattern=north east lines,pattern color=lyyblue!60] coordinates {(Supported,69) (Refuted,8) (NEI,23)};
          \addplot [draw=lyygreen!60,fill=lyygreen!60,pattern=horizontal lines,pattern color=lyygreen!60] coordinates {(Supported,79) (Refuted,14) (NEI,7)};
        \nextgroupplot[
          width=0.25\linewidth,
          symbolic x coords={Explicit,Implicit},
          enlarge x limits=0.6,
          enlarge y limits={upper,value=0.3},
          ylabel={\#Examples},
          ylabel near ticks,
          ybar=0pt,
          xtick=data,
          ytick=\empty,
          ymin=0,
          nodes near coords,
          every node near coord/.append style={rotate=90,anchor=west,font=\scriptsize},
          every tick label/.append style={font=\footnotesize},
          xticklabel style={rotate=45,anchor=north east,font=\footnotesize,inner sep=0pt,outer sep=2pt},
          bar width=7pt,
          xmajorgrids=true,
          ymajorgrids=true,
          grid style=dashed,
          label style={font=\small},
        ]
          \addplot [draw=lyyblue!60,fill=lyyblue!60,pattern=north east lines,pattern color=lyyblue!60] coordinates {(Explicit,29) (Implicit,40)};
          \addplot [draw=lyygreen!60,fill=lyygreen!60,pattern=horizontal lines,pattern color=lyygreen!60] coordinates {(Explicit,27) (Implicit,52)};
        \nextgroupplot[
          width=0.25\linewidth,
          symbolic x coords={Reasonable,Unreasonable,Hard},
          enlarge x limits=0.3,
          enlarge y limits={upper,value=0.3},
          ylabel={\#Examples},
          ylabel near ticks,
          ybar=0pt,
          xtick=data,
          ytick=\empty,
          ymin=0,
          nodes near coords,
          every node near coord/.append style={rotate=90,anchor=west,font=\scriptsize},
          every tick label/.append style={font=\footnotesize},
          xticklabel style={rotate=45,anchor=north east,font=\footnotesize,inner sep=0pt,outer sep=2pt},
          bar width=7pt,
          xmajorgrids=true,
          ymajorgrids=true,
          grid style=dashed,
          label style={font=\small},
        ]
          \addplot [draw=lyyblue!60,fill=lyyblue!60,pattern=north east lines,pattern color=lyyblue!60] coordinates {(Reasonable,9) (Unreasonable,0) (Hard,14)};
          \addplot [draw=lyygreen!60,fill=lyygreen!60,pattern=horizontal lines,pattern color=lyygreen!60] coordinates {(Reasonable,4) (Unreasonable,0) (Hard,3)};
        \end{groupplot}
    \end{tikzpicture}
    \caption{The human evaluation results of generated knowledge from fine-tuned DialoGPT-large on WoW seen and unseen test sets.}
    \label{fig:knowledge}
    \vspace{-0.4cm}
\end{figure*}

%% file: figures/knowledge-t5.tex
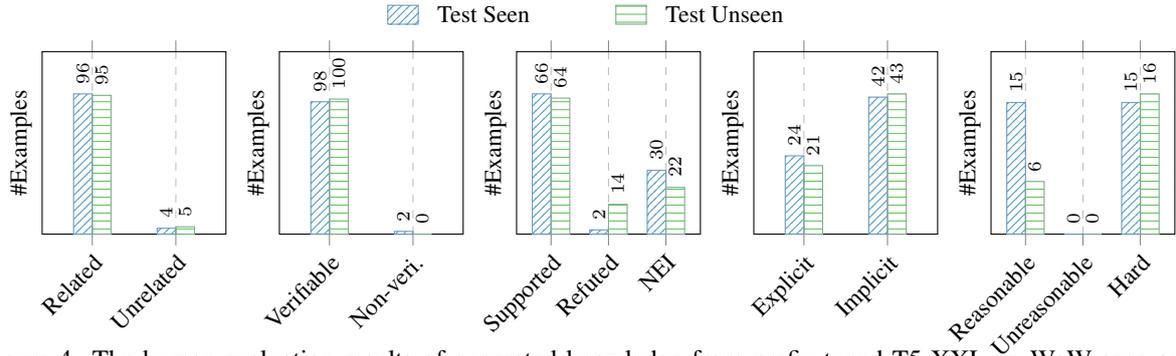
\begin{figure*}[t!]
    \centering
    \makeatletter
    \let\ref\@refstar
    \ref{grouplegend2}
    \makeatother
    \vspace{-0.2in}
    \begin{tikzpicture}
        \begin{groupplot}[
            group style={group size=5 by 1, horizontal sep=20pt},
            width=1.0\textwidth,
            height=4cm,
            legend cell align={left},
            legend pos=north west,
            enlargelimits=0.1,
            legend style={
                font=\small,
                draw=none,
                column sep=5pt,
                /tikz/every even column/.append style={column sep=30pt},
                legend columns=2,
            },
        ]
        \nextgroupplot[
          width=0.25\linewidth,
          symbolic x coords={Related,Unrelated},
          enlarge x limits=0.6,
          enlarge y limits={upper,value=0.3},
          ylabel={\#Examples},
          ylabel near ticks,
          ybar=0pt,
          xtick=data,
          ytick=\empty,
          ymin=0,
          nodes near coords,
          every node near coord/.append style={rotate=90,anchor=west,font=\scriptsize},
          every tick label/.append style={font=\footnotesize},
          xticklabel style={rotate=45,anchor=north east,font=\footnotesize,inner sep=0pt,outer sep=2pt},
          bar width=7pt,
          xmajorgrids=true,
          ymajorgrids=true,
          grid style=dashed,
          label style={font=\small},
          legend to name=grouplegend2,
          legend image code/.code={\draw [#1] (0cm,-0.1cm) rectangle ++(0.4cm,0.25cm);},
        ]
          \addplot [draw=lyyblue!60,fill=lyyblue!60,pattern=north east lines,pattern color=lyyblue!60] coordinates {(Related,96) (Unrelated,4)};
          \addlegendentry{Test Seen}
          \addplot [draw=lyygreen!60,fill=lyygreen!60,pattern=horizontal lines,pattern color=lyygreen!60] coordinates {(Related,95) (Unrelated,5)};
          \addlegendentry{Test Unseen}
        \nextgroupplot[
          width=0.25\linewidth,
          symbolic x coords={Verifiable,Non-veri.},
          enlarge x limits=0.6,
          enlarge y limits={upper,value=0.35},
          ylabel={\#Examples},
          ylabel near ticks,
          ybar=0pt,
          xtick=data,
          ytick=\empty,
          ymin=0,
          nodes near coords,
          every node near coord/.append style={rotate=90,anchor=west,font=\scriptsize},
          every tick label/.append style={font=\footnotesize},
          xticklabel style={rotate=45,anchor=north east,font=\footnotesize,inner sep=0pt,outer sep=2pt},
          bar width=7pt,
          xmajorgrids=true,
          ymajorgrids=true,
          grid style=dashed,
          label style={font=\small},
        ]
          \addplot [draw=lyyblue!60,fill=lyyblue!60,pattern=north east lines,pattern color=lyyblue!60] coordinates {(Verifiable,98) (Non-veri.,2)};
          \addplot [draw=lyygreen!60,fill=lyygreen!60,pattern=horizontal lines,pattern color=lyygreen!60] coordinates {(Verifiable,100) (Non-veri.,0)};
        \nextgroupplot[
          width=0.25\linewidth,
          symbolic x coords={Supported,Refuted,NEI},
          enlarge x limits=0.3,
          enlarge y limits={upper,value=0.3},
          ylabel={\#Examples},
          ylabel near ticks,
          ybar=0pt,
          xtick=data,
          ytick=\empty,
          ymin=0,
          nodes near coords,
          every node near coord/.append style={rotate=90,anchor=west,font=\scriptsize},
          every tick label/.append style={font=\footnotesize},
          xticklabel style={rotate=45,anchor=north east,font=\footnotesize,inner sep=0pt,outer sep=2pt},
          bar width=7pt,
          xmajorgrids=true,
          ymajorgrids=true,
          grid style=dashed,
          label style={font=\small},
        ]
          \addplot [draw=lyyblue!60,fill=lyyblue!60,pattern=north east lines,pattern color=lyyblue!60] coordinates {(Supported,66) (Refuted,2) (NEI,30)};
          \addplot [draw=lyygreen!60,fill=lyygreen!60,pattern=horizontal lines,pattern color=lyygreen!60] coordinates {(Supported,64) (Refuted,14) (NEI,22)};
        \nextgroupplot[
          width=0.25\linewidth,
          symbolic x coords={Explicit,Implicit},
          enlarge x limits=0.6,
          enlarge y limits={upper,value=0.3},
          ylabel={\#Examples},
          ylabel near ticks,
          ybar=0pt,
          xtick=data,
          ytick=\empty,
          ymin=0,
          nodes near coords,
          every node near coord/.append style={rotate=90,anchor=west,font=\scriptsize},
          every tick label/.append style={font=\footnotesize},
          xticklabel style={rotate=45,anchor=north east,font=\footnotesize,inner sep=0pt,outer sep=2pt},
          bar width=7pt,
          xmajorgrids=true,
          ymajorgrids=true,
          grid style=dashed,
          label style={font=\small},
        ]
          \addplot [draw=lyyblue!60,fill=lyyblue!60,pattern=north east lines,pattern color=lyyblue!60] coordinates {(Explicit,24) (Implicit,42)};
          \addplot [draw=lyygreen!60,fill=lyygreen!60,pattern=horizontal lines,pattern color=lyygreen!60] coordinates {(Explicit,21) (Implicit,43)};
        \nextgroupplot[
          width=0.25\linewidth,
          symbolic x coords={Reasonable,Unreasonable,Hard},
          enlarge x limits=0.3,
          enlarge y limits={upper,value=0.3},
          ylabel={\#Examples},
          ylabel near ticks,
          ybar=0pt,
          xtick=data,
          ytick=\empty,
          ymin=0,
          nodes near coords,
          every node near coord/.append style={rotate=90,anchor=west,font=\scriptsize},
          every tick label/.append style={font=\footnotesize},
          xticklabel style={rotate=45,anchor=north east,font=\footnotesize,inner sep=0pt,outer sep=2pt},
          bar width=7pt,
          xmajorgrids=true,
          ymajorgrids=true,
          grid style=dashed,
          label style={font=\small},
        ]
          \addplot [draw=lyyblue!60,fill=lyyblue!60,pattern=north east lines,pattern color=lyyblue!60] coordinates {(Reasonable,15) (Unreasonable,0) (Hard,15)};
          \addplot [draw=lyygreen!60,fill=lyygreen!60,pattern=horizontal lines,pattern color=lyygreen!60] coordinates {(Reasonable,6) (Unreasonable,0) (Hard,16)};
        \end{groupplot}
    \end{tikzpicture}
    \caption{The human evaluation results of generated knowledge from prefix-tuned T5-XXL on WoW seen and unseen test sets.}
    \label{fig:knowledge-t5}
    \vspace{-0.4cm}
\end{figure*}

%% file: emnlp2022.bbl
\begin{thebibliography}{44}
\expandafter\ifx\csname natexlab\endcsname\relax\def\natexlab#1{#1}\fi

\bibitem[{Devlin et~al.(2019)Devlin, Chang, Lee, and
  Toutanova}]{devlin-etal-2019-bert}
Jacob Devlin, Ming-Wei Chang, Kenton Lee, and Kristina Toutanova. 2019.
\newblock \href {https://doi.org/10.18653/v1/N19-1423} {{BERT}: Pre-training of
  deep bidirectional transformers for language understanding}.
\newblock In \emph{Proceedings of the 2019 Conference of the North {A}merican
  Chapter of the Association for Computational Linguistics: Human Language
  Technologies, Volume 1 (Long and Short Papers)}, pages 4171--4186,
  Minneapolis, Minnesota. Association for Computational Linguistics.

\bibitem[{Dinan et~al.(2019)Dinan, Roller, Shuster, Fan, Auli, and
  Weston}]{DBLP:conf/iclr/DinanRSFAW19}
Emily Dinan, Stephen Roller, Kurt Shuster, Angela Fan, Michael Auli, and Jason
  Weston. 2019.
\newblock \href {https://openreview.net/forum?id=r1l73iRqKm} {Wizard of
  wikipedia: Knowledge-powered conversational agents}.
\newblock In \emph{7th International Conference on Learning Representations,
  {ICLR} 2019, New Orleans, LA, USA, May 6-9, 2019}. OpenReview.net.

\bibitem[{Dziri et~al.(2021)Dziri, Madotto, Za{\"\i}ane, and
  Bose}]{dziri-etal-2021-neural}
Nouha Dziri, Andrea Madotto, Osmar Za{\"\i}ane, and Avishek~Joey Bose. 2021.
\newblock \href {https://doi.org/10.18653/v1/2021.emnlp-main.168} {Neural path
  hunter: Reducing hallucination in dialogue systems via path grounding}.
\newblock In \emph{Proceedings of the 2021 Conference on Empirical Methods in
  Natural Language Processing}, pages 2197--2214, Online and Punta Cana,
  Dominican Republic. Association for Computational Linguistics.

\bibitem[{Fan et~al.(2018)Fan, Lewis, and Dauphin}]{fan-etal-2018-hierarchical}
Angela Fan, Mike Lewis, and Yann Dauphin. 2018.
\newblock \href {https://doi.org/10.18653/v1/P18-1082} {Hierarchical neural
  story generation}.
\newblock In \emph{Proceedings of the 56th Annual Meeting of the Association
  for Computational Linguistics (Volume 1: Long Papers)}, pages 889--898,
  Melbourne, Australia. Association for Computational Linguistics.

\bibitem[{Fleiss(1971)}]{fleiss1971measuring}
Joseph~L Fleiss. 1971.
\newblock Measuring nominal scale agreement among many raters.
\newblock \emph{Psychological bulletin}, 76(5):378.

\bibitem[{Goodfellow et~al.(2013)Goodfellow, Mirza, Xiao, Courville, and
  Bengio}]{goodfellow2013empirical}
Ian~J Goodfellow, Mehdi Mirza, Da~Xiao, Aaron Courville, and Yoshua Bengio.
  2013.
\newblock An empirical investigation of catastrophic forgetting in
  gradient-based neural networks.
\newblock \emph{arXiv preprint arXiv:1312.6211}.

\bibitem[{Gopalakrishnan et~al.(2019)Gopalakrishnan, Hedayatnia, Chen,
  Gottardi, Kwatra, Venkatesh, Gabriel, and
  Hakkani-Tür}]{gopalakrishnan19_interspeech}
Karthik Gopalakrishnan, Behnam Hedayatnia, Qinlang Chen, Anna Gottardi, Sanjeev
  Kwatra, Anu Venkatesh, Raefer Gabriel, and Dilek Hakkani-Tür. 2019.
\newblock \href {https://doi.org/10.21437/Interspeech.2019-3079}
  {{Topical-Chat: Towards Knowledge-Grounded Open-Domain Conversations}}.
\newblock In \emph{Proc. Interspeech 2019}, pages 1891--1895.

\bibitem[{Gupta et~al.(2022)Gupta, Wu, Liu, and
  Xiong}]{gupta-etal-2022-dialfact}
Prakhar Gupta, Chien-Sheng Wu, Wenhao Liu, and Caiming Xiong. 2022.
\newblock \href {https://doi.org/10.18653/v1/2022.acl-long.263} {{D}ial{F}act:
  A benchmark for fact-checking in dialogue}.
\newblock In \emph{Proceedings of the 60th Annual Meeting of the Association
  for Computational Linguistics (Volume 1: Long Papers)}, pages 3785--3801,
  Dublin, Ireland. Association for Computational Linguistics.

\bibitem[{Hinton et~al.(2015)Hinton, Vinyals, and
  Dean}]{DBLP:journals/corr/HintonVD15}
Geoffrey~E. Hinton, Oriol Vinyals, and Jeffrey Dean. 2015.
\newblock \href {http://arxiv.org/abs/1503.02531} {Distilling the knowledge in
  a neural network}.
\newblock \emph{CoRR}, abs/1503.02531.

\bibitem[{Huang et~al.(2021)Huang, He, Bao, Wang, Wu, and
  Wang}]{huang-etal-2021-plato}
Xinxian Huang, Huang He, Siqi Bao, Fan Wang, Hua Wu, and Haifeng Wang. 2021.
\newblock \href {https://doi.org/10.18653/v1/2021.nlp4convai-1.14}
  {{PLATO}-{KAG}: Unsupervised knowledge-grounded conversation via joint
  modeling}.
\newblock In \emph{Proceedings of the 3rd Workshop on Natural Language
  Processing for Conversational AI}, pages 143--154, Online. Association for
  Computational Linguistics.

\bibitem[{Karpukhin et~al.(2020)Karpukhin, Oguz, Min, Lewis, Wu, Edunov, Chen,
  and Yih}]{karpukhin-etal-2020-dense}
Vladimir Karpukhin, Barlas Oguz, Sewon Min, Patrick Lewis, Ledell Wu, Sergey
  Edunov, Danqi Chen, and Wen-tau Yih. 2020.
\newblock \href {https://doi.org/10.18653/v1/2020.emnlp-main.550} {Dense
  passage retrieval for open-domain question answering}.
\newblock In \emph{Proceedings of the 2020 Conference on Empirical Methods in
  Natural Language Processing (EMNLP)}, pages 6769--6781, Online. Association
  for Computational Linguistics.

\bibitem[{Kim et~al.(2020)Kim, Ahn, and Kim}]{DBLP:conf/iclr/KimAK20}
Byeongchang Kim, Jaewoo Ahn, and Gunhee Kim. 2020.
\newblock \href {https://openreview.net/forum?id=Hke0K1HKwr} {Sequential latent
  knowledge selection for knowledge-grounded dialogue}.
\newblock In \emph{8th International Conference on Learning Representations,
  {ICLR} 2020, Addis Ababa, Ethiopia, April 26-30, 2020}. OpenReview.net.

\bibitem[{Kingma and Welling(2014)}]{DBLP:journals/corr/KingmaW13}
Diederik~P. Kingma and Max Welling. 2014.
\newblock \href {http://arxiv.org/abs/1312.6114} {Auto-encoding variational
  bayes}.
\newblock In \emph{2nd International Conference on Learning Representations,
  {ICLR} 2014, Banff, AB, Canada, April 14-16, 2014, Conference Track
  Proceedings}.

\bibitem[{Komeili et~al.(2022)Komeili, Shuster, and
  Weston}]{komeili-etal-2022-internet}
Mojtaba Komeili, Kurt Shuster, and Jason Weston. 2022.
\newblock \href {https://doi.org/10.18653/v1/2022.acl-long.579}
  {{I}nternet-augmented dialogue generation}.
\newblock In \emph{Proceedings of the 60th Annual Meeting of the Association
  for Computational Linguistics (Volume 1: Long Papers)}, pages 8460--8478,
  Dublin, Ireland. Association for Computational Linguistics.

\bibitem[{Kool et~al.(2019)Kool, van Hoof, and
  Welling}]{DBLP:conf/icml/KoolHW19}
Wouter Kool, Herke van Hoof, and Max Welling. 2019.
\newblock \href {http://proceedings.mlr.press/v97/kool19a.html} {Stochastic
  beams and where to find them: The gumbel-top-k trick for sampling sequences
  without replacement}.
\newblock In \emph{Proceedings of the 36th International Conference on Machine
  Learning, {ICML} 2019, 9-15 June 2019, Long Beach, California, {USA}},
  volume~97 of \emph{Proceedings of Machine Learning Research}, pages
  3499--3508. {PMLR}.

\bibitem[{Lester et~al.(2021)Lester, Al-Rfou, and
  Constant}]{lester-etal-2021-power}
Brian Lester, Rami Al-Rfou, and Noah Constant. 2021.
\newblock \href {https://doi.org/10.18653/v1/2021.emnlp-main.243} {The power of
  scale for parameter-efficient prompt tuning}.
\newblock In \emph{Proceedings of the 2021 Conference on Empirical Methods in
  Natural Language Processing}, pages 3045--3059, Online and Punta Cana,
  Dominican Republic. Association for Computational Linguistics.

\bibitem[{Lewis et~al.(2021)Lewis, Stenetorp, and
  Riedel}]{lewis-etal-2021-question}
Patrick Lewis, Pontus Stenetorp, and Sebastian Riedel. 2021.
\newblock \href {https://doi.org/10.18653/v1/2021.eacl-main.86} {Question and
  answer test-train overlap in open-domain question answering datasets}.
\newblock In \emph{Proceedings of the 16th Conference of the European Chapter
  of the Association for Computational Linguistics: Main Volume}, pages
  1000--1008, Online. Association for Computational Linguistics.

\bibitem[{Li et~al.(2016)Li, Galley, Brockett, Gao, and
  Dolan}]{li-etal-2016-diversity}
Jiwei Li, Michel Galley, Chris Brockett, Jianfeng Gao, and Bill Dolan. 2016.
\newblock \href {https://doi.org/10.18653/v1/N16-1014} {A diversity-promoting
  objective function for neural conversation models}.
\newblock In \emph{Proceedings of the 2016 Conference of the North {A}merican
  Chapter of the Association for Computational Linguistics: Human Language
  Technologies}, pages 110--119, San Diego, California. Association for
  Computational Linguistics.

\bibitem[{Li and Liang(2021)}]{li-liang-2021-prefix}
Xiang~Lisa Li and Percy Liang. 2021.
\newblock \href {https://doi.org/10.18653/v1/2021.acl-long.353} {Prefix-tuning:
  Optimizing continuous prompts for generation}.
\newblock In \emph{Proceedings of the 59th Annual Meeting of the Association
  for Computational Linguistics and the 11th International Joint Conference on
  Natural Language Processing (Volume 1: Long Papers)}, pages 4582--4597,
  Online. Association for Computational Linguistics.

\bibitem[{Lian et~al.(2019)Lian, Xie, Wang, Peng, and
  Wu}]{DBLP:conf/ijcai/LianXWPW19}
Rongzhong Lian, Min Xie, Fan Wang, Jinhua Peng, and Hua Wu. 2019.
\newblock \href {https://doi.org/10.24963/ijcai.2019/706} {Learning to select
  knowledge for response generation in dialog systems}.
\newblock In \emph{Proceedings of the Twenty-Eighth International Joint
  Conference on Artificial Intelligence, {IJCAI} 2019, Macao, China, August
  10-16, 2019}, pages 5081--5087. ijcai.org.

\bibitem[{Lin et~al.(2021)Lin, Li, Wang, Li, Du, Xiao, and
  Zhu}]{lin-etal-2021-weight}
Ye~Lin, Yanyang Li, Ziyang Wang, Bei Li, Quan Du, Tong Xiao, and Jingbo Zhu.
  2021.
\newblock \href {https://doi.org/10.18653/v1/2021.acl-long.162} {Weight
  distillation: Transferring the knowledge in neural network parameters}.
\newblock In \emph{Proceedings of the 59th Annual Meeting of the Association
  for Computational Linguistics and the 11th International Joint Conference on
  Natural Language Processing (Volume 1: Long Papers)}, pages 2076--2088,
  Online. Association for Computational Linguistics.

\bibitem[{Liu et~al.(2022{\natexlab{a}})Liu, Liu, Lu, Welleck, West, Le~Bras,
  Choi, and Hajishirzi}]{liu-etal-2022-generated}
Jiacheng Liu, Alisa Liu, Ximing Lu, Sean Welleck, Peter West, Ronan Le~Bras,
  Yejin Choi, and Hannaneh Hajishirzi. 2022{\natexlab{a}}.
\newblock \href {https://doi.org/10.18653/v1/2022.acl-long.225} {Generated
  knowledge prompting for commonsense reasoning}.
\newblock In \emph{Proceedings of the 60th Annual Meeting of the Association
  for Computational Linguistics (Volume 1: Long Papers)}, pages 3154--3169,
  Dublin, Ireland. Association for Computational Linguistics.

\bibitem[{Liu et~al.(2022{\natexlab{b}})Liu, Patwary, Prenger, Prabhumoye,
  Ping, Shoeybi, and Catanzaro}]{liu-etal-2022-multi}
Zihan Liu, Mostofa Patwary, Ryan Prenger, Shrimai Prabhumoye, Wei Ping,
  Mohammad Shoeybi, and Bryan Catanzaro. 2022{\natexlab{b}}.
\newblock \href {https://doi.org/10.18653/v1/2022.findings-acl.104}
  {Multi-stage prompting for knowledgeable dialogue generation}.
\newblock In \emph{Findings of the Association for Computational Linguistics:
  ACL 2022}, pages 1317--1337, Dublin, Ireland. Association for Computational
  Linguistics.

\bibitem[{Maynez et~al.(2020)Maynez, Narayan, Bohnet, and
  McDonald}]{maynez-etal-2020-faithfulness}
Joshua Maynez, Shashi Narayan, Bernd Bohnet, and Ryan McDonald. 2020.
\newblock \href {https://doi.org/10.18653/v1/2020.acl-main.173} {On
  faithfulness and factuality in abstractive summarization}.
\newblock In \emph{Proceedings of the 58th Annual Meeting of the Association
  for Computational Linguistics}, pages 1906--1919, Online. Association for
  Computational Linguistics.

\bibitem[{Miller et~al.(2017)Miller, Feng, Batra, Bordes, Fisch, Lu, Parikh,
  and Weston}]{miller-etal-2017-parlai}
Alexander Miller, Will Feng, Dhruv Batra, Antoine Bordes, Adam Fisch, Jiasen
  Lu, Devi Parikh, and Jason Weston. 2017.
\newblock \href {https://doi.org/10.18653/v1/D17-2014} {{P}arl{AI}: A dialog
  research software platform}.
\newblock In \emph{Proceedings of the 2017 Conference on Empirical Methods in
  Natural Language Processing: System Demonstrations}, pages 79--84,
  Copenhagen, Denmark. Association for Computational Linguistics.

\bibitem[{Moghe et~al.(2018)Moghe, Arora, Banerjee, and
  Khapra}]{moghe-etal-2018-towards}
Nikita Moghe, Siddhartha Arora, Suman Banerjee, and Mitesh~M. Khapra. 2018.
\newblock \href {https://doi.org/10.18653/v1/D18-1255} {Towards exploiting
  background knowledge for building conversation systems}.
\newblock In \emph{Proceedings of the 2018 Conference on Empirical Methods in
  Natural Language Processing}, pages 2322--2332, Brussels, Belgium.
  Association for Computational Linguistics.

\bibitem[{Petroni et~al.(2019)Petroni, Rockt{\"a}schel, Riedel, Lewis, Bakhtin,
  Wu, and Miller}]{petroni-etal-2019-language}
Fabio Petroni, Tim Rockt{\"a}schel, Sebastian Riedel, Patrick Lewis, Anton
  Bakhtin, Yuxiang Wu, and Alexander Miller. 2019.
\newblock \href {https://doi.org/10.18653/v1/D19-1250} {Language models as
  knowledge bases?}
\newblock In \emph{Proceedings of the 2019 Conference on Empirical Methods in
  Natural Language Processing and the 9th International Joint Conference on
  Natural Language Processing (EMNLP-IJCNLP)}, pages 2463--2473, Hong Kong,
  China. Association for Computational Linguistics.

\bibitem[{Radford et~al.(2019)Radford, Wu, Child, Luan, Amodei, Sutskever
  et~al.}]{radford2019language}
Alec Radford, Jeffrey Wu, Rewon Child, David Luan, Dario Amodei, Ilya
  Sutskever, et~al. 2019.
\newblock Language models are unsupervised multitask learners.
\newblock \emph{OpenAI blog}, 1(8):9.

\bibitem[{Raffel et~al.(2020)Raffel, Shazeer, Roberts, Lee, Narang, Matena,
  Zhou, Li, and Liu}]{JMLR:v21:20-074}
Colin Raffel, Noam Shazeer, Adam Roberts, Katherine Lee, Sharan Narang, Michael
  Matena, Yanqi Zhou, Wei Li, and Peter~J. Liu. 2020.
\newblock \href {http://jmlr.org/papers/v21/20-074.html} {Exploring the limits
  of transfer learning with a unified text-to-text transformer}.
\newblock \emph{Journal of Machine Learning Research}, 21(140):1--67.

\bibitem[{Roberts et~al.(2020)Roberts, Raffel, and
  Shazeer}]{roberts-etal-2020-much}
Adam Roberts, Colin Raffel, and Noam Shazeer. 2020.
\newblock \href {https://doi.org/10.18653/v1/2020.emnlp-main.437} {How much
  knowledge can you pack into the parameters of a language model?}
\newblock In \emph{Proceedings of the 2020 Conference on Empirical Methods in
  Natural Language Processing (EMNLP)}, pages 5418--5426, Online. Association
  for Computational Linguistics.

\bibitem[{Romero et~al.(2015)Romero, Ballas, Kahou, Chassang, Gatta, and
  Bengio}]{DBLP:journals/corr/RomeroBKCGB14}
Adriana Romero, Nicolas Ballas, Samira~Ebrahimi Kahou, Antoine Chassang, Carlo
  Gatta, and Yoshua Bengio. 2015.
\newblock \href {http://arxiv.org/abs/1412.6550} {Fitnets: Hints for thin deep
  nets}.
\newblock In \emph{3rd International Conference on Learning Representations,
  {ICLR} 2015, San Diego, CA, USA, May 7-9, 2015, Conference Track
  Proceedings}.

\bibitem[{Shuster et~al.(2021)Shuster, Poff, Chen, Kiela, and
  Weston}]{shuster-etal-2021-retrieval-augmentation}
Kurt Shuster, Spencer Poff, Moya Chen, Douwe Kiela, and Jason Weston. 2021.
\newblock \href {https://doi.org/10.18653/v1/2021.findings-emnlp.320}
  {Retrieval augmentation reduces hallucination in conversation}.
\newblock In \emph{Findings of the Association for Computational Linguistics:
  EMNLP 2021}, pages 3784--3803, Punta Cana, Dominican Republic. Association
  for Computational Linguistics.

\bibitem[{Thoppilan et~al.(2022)Thoppilan, Freitas, Hall, Shazeer,
  Kulshreshtha, Cheng, Jin, Bos, Baker, Du, Li, Lee, Zheng, Ghafouri, Menegali,
  Huang, Krikun, Lepikhin, Qin, Chen, Xu, Chen, Roberts, Bosma, Zhou, Chang,
  Krivokon, Rusch, Pickett, Meier{-}Hellstern, Morris, Doshi, Santos, Duke,
  Soraker, Zevenbergen, Prabhakaran, Diaz, Hutchinson, Olson, Molina,
  Hoffman{-}John, Lee, Aroyo, Rajakumar, Butryna, Lamm, Kuzmina, Fenton, Cohen,
  Bernstein, Kurzweil, Aguera{-}Arcas, Cui, Croak, Chi, and
  Le}]{DBLP:journals/corr/abs-2201-08239}
Romal Thoppilan, Daniel~De Freitas, Jamie Hall, Noam Shazeer, Apoorv
  Kulshreshtha, Heng{-}Tze Cheng, Alicia Jin, Taylor Bos, Leslie Baker, Yu~Du,
  YaGuang Li, Hongrae Lee, Huaixiu~Steven Zheng, Amin Ghafouri, Marcelo
  Menegali, Yanping Huang, Maxim Krikun, Dmitry Lepikhin, James Qin, Dehao
  Chen, Yuanzhong Xu, Zhifeng Chen, Adam Roberts, Maarten Bosma, Yanqi Zhou,
  Chung{-}Ching Chang, Igor Krivokon, Will Rusch, Marc Pickett, Kathleen~S.
  Meier{-}Hellstern, Meredith~Ringel Morris, Tulsee Doshi, Renelito~Delos
  Santos, Toju Duke, Johnny Soraker, Ben Zevenbergen, Vinodkumar Prabhakaran,
  Mark Diaz, Ben Hutchinson, Kristen Olson, Alejandra Molina, Erin
  Hoffman{-}John, Josh Lee, Lora Aroyo, Ravi Rajakumar, Alena Butryna, Matthew
  Lamm, Viktoriya Kuzmina, Joe Fenton, Aaron Cohen, Rachel Bernstein, Ray
  Kurzweil, Blaise Aguera{-}Arcas, Claire Cui, Marian Croak, Ed~H. Chi, and
  Quoc Le. 2022.
\newblock \href {http://arxiv.org/abs/2201.08239} {Lamda: Language models for
  dialog applications}.
\newblock \emph{CoRR}, abs/2201.08239.

\bibitem[{Vaswani et~al.(2017)Vaswani, Shazeer, Parmar, Uszkoreit, Jones,
  Gomez, Kaiser, and Polosukhin}]{DBLP:conf/nips/VaswaniSPUJGKP17}
Ashish Vaswani, Noam Shazeer, Niki Parmar, Jakob Uszkoreit, Llion Jones,
  Aidan~N. Gomez, Lukasz Kaiser, and Illia Polosukhin. 2017.
\newblock \href
  {https://proceedings.neurips.cc/paper/2017/hash/3f5ee243547dee91fbd053c1c4a845aa-Abstract.html}
  {Attention is all you need}.
\newblock In \emph{Advances in Neural Information Processing Systems 30: Annual
  Conference on Neural Information Processing Systems 2017, December 4-9, 2017,
  Long Beach, CA, {USA}}, pages 5998--6008.

\bibitem[{Wang et~al.(2021)Wang, Liu, and Zhang}]{wang-etal-2021-generative}
Cunxiang Wang, Pai Liu, and Yue Zhang. 2021.
\newblock \href {https://doi.org/10.18653/v1/2021.acl-long.251} {Can generative
  pre-trained language models serve as knowledge bases for closed-book {QA}?}
\newblock In \emph{Proceedings of the 59th Annual Meeting of the Association
  for Computational Linguistics and the 11th International Joint Conference on
  Natural Language Processing (Volume 1: Long Papers)}, pages 3241--3251,
  Online. Association for Computational Linguistics.

\bibitem[{Wang et~al.(2022)Wang, Qin, Hui, Li, Yang, Wang, Li, Sun, Huang, Si,
  and Li}]{DBLP:conf/kdd/WangQHLYWLSHSL22}
Lihan Wang, Bowen Qin, Binyuan Hui, Bowen Li, Min Yang, Bailin Wang, Binhua Li,
  Jian Sun, Fei Huang, Luo Si, and Yongbin Li. 2022.
\newblock \href {https://doi.org/10.1145/3534678.3539305} {Proton: Probing
  schema linking information from pre-trained language models for text-to-sql
  parsing}.
\newblock In \emph{{KDD} '22: The 28th {ACM} {SIGKDD} Conference on Knowledge
  Discovery and Data Mining, Washington, DC, USA, August 14 - 18, 2022}, pages
  1889--1898. {ACM}.

\bibitem[{Wang et~al.(2020)Wang, Wei, Dong, Bao, Yang, and
  Zhou}]{DBLP:conf/nips/WangW0B0020}
Wenhui Wang, Furu Wei, Li~Dong, Hangbo Bao, Nan Yang, and Ming Zhou. 2020.
\newblock \href
  {https://proceedings.neurips.cc/paper/2020/hash/3f5ee243547dee91fbd053c1c4a845aa-Abstract.html}
  {Minilm: Deep self-attention distillation for task-agnostic compression of
  pre-trained transformers}.
\newblock In \emph{Advances in Neural Information Processing Systems 33: Annual
  Conference on Neural Information Processing Systems 2020, NeurIPS 2020,
  December 6-12, 2020, virtual}.

\bibitem[{Wei et~al.(2022)Wei, Wang, Schuurmans, Bosma, Chi, Le, and
  Zhou}]{DBLP:journals/corr/abs-2201-11903}
Jason Wei, Xuezhi Wang, Dale Schuurmans, Maarten Bosma, Ed~H. Chi, Quoc Le, and
  Denny Zhou. 2022.
\newblock \href {http://arxiv.org/abs/2201.11903} {Chain of thought prompting
  elicits reasoning in large language models}.
\newblock \emph{CoRR}, abs/2201.11903.

\bibitem[{Yang et~al.(2018)Yang, Huang, and Ma}]{yang-etal-2018-breaking}
Yilin Yang, Liang Huang, and Mingbo Ma. 2018.
\newblock \href {https://doi.org/10.18653/v1/D18-1342} {Breaking the beam
  search curse: A study of (re-)scoring methods and stopping criteria for
  neural machine translation}.
\newblock In \emph{Proceedings of the 2018 Conference on Empirical Methods in
  Natural Language Processing}, pages 3054--3059, Brussels, Belgium.
  Association for Computational Linguistics.

\bibitem[{Zhang et~al.(2018)Zhang, Dinan, Urbanek, Szlam, Kiela, and
  Weston}]{zhang-etal-2018-personalizing}
Saizheng Zhang, Emily Dinan, Jack Urbanek, Arthur Szlam, Douwe Kiela, and Jason
  Weston. 2018.
\newblock \href {https://doi.org/10.18653/v1/P18-1205} {Personalizing dialogue
  agents: {I} have a dog, do you have pets too?}
\newblock In \emph{Proceedings of the 56th Annual Meeting of the Association
  for Computational Linguistics (Volume 1: Long Papers)}, pages 2204--2213,
  Melbourne, Australia. Association for Computational Linguistics.

\bibitem[{Zhang et~al.(2021)Zhang, Wu, Katiyar, Weinberger, and
  Artzi}]{DBLP:conf/iclr/0007WKWA21}
Tianyi Zhang, Felix Wu, Arzoo Katiyar, Kilian~Q. Weinberger, and Yoav Artzi.
  2021.
\newblock \href {https://openreview.net/forum?id=cO1IH43yUF} {Revisiting
  few-sample {BERT} fine-tuning}.
\newblock In \emph{9th International Conference on Learning Representations,
  {ICLR} 2021, Virtual Event, Austria, May 3-7, 2021}. OpenReview.net.

\bibitem[{Zhang et~al.(2020)Zhang, Sun, Galley, Chen, Brockett, Gao, Gao, Liu,
  and Dolan}]{zhang-etal-2020-dialogpt}
Yizhe Zhang, Siqi Sun, Michel Galley, Yen-Chun Chen, Chris Brockett, Xiang Gao,
  Jianfeng Gao, Jingjing Liu, and Bill Dolan. 2020.
\newblock \href {https://doi.org/10.18653/v1/2020.acl-demos.30} {{DIALOGPT} :
  Large-scale generative pre-training for conversational response generation}.
\newblock In \emph{Proceedings of the 58th Annual Meeting of the Association
  for Computational Linguistics: System Demonstrations}, pages 270--278,
  Online. Association for Computational Linguistics.

\bibitem[{Zhao et~al.(2020)Zhao, Wu, Xu, Tao, Zhao, and
  Yan}]{zhao-etal-2020-knowledge-grounded}
Xueliang Zhao, Wei Wu, Can Xu, Chongyang Tao, Dongyan Zhao, and Rui Yan. 2020.
\newblock \href {https://doi.org/10.18653/v1/2020.emnlp-main.272}
  {Knowledge-grounded dialogue generation with pre-trained language models}.
\newblock In \emph{Proceedings of the 2020 Conference on Empirical Methods in
  Natural Language Processing (EMNLP)}, pages 3377--3390, Online. Association
  for Computational Linguistics.

\bibitem[{Zhou et~al.(2018)Zhou, Prabhumoye, and
  Black}]{zhou-etal-2018-dataset}
Kangyan Zhou, Shrimai Prabhumoye, and Alan~W Black. 2018.
\newblock \href {https://doi.org/10.18653/v1/D18-1076} {A dataset for document
  grounded conversations}.
\newblock In \emph{Proceedings of the 2018 Conference on Empirical Methods in
  Natural Language Processing}, pages 708--713, Brussels, Belgium. Association
  for Computational Linguistics.

\end{thebibliography}
